\documentclass[conference]{IEEEtran}
\IEEEoverridecommandlockouts
\usepackage{cite}
\usepackage{amsmath,amssymb,amsfonts}
\usepackage{algorithmic}
\usepackage{graphicx}
\usepackage{textcomp}
\usepackage{xcolor}
\usepackage{subcaption}

\usepackage{graphicx}
\usepackage{booktabs}
\usepackage{longtable}
\usepackage{tabulary}
\usepackage{hhline}
\usepackage{multirow, rotating}
\usepackage{threeparttable}

\newcommand{\tcen}[1]{\multicolumn{1}{c}{#1}}
\def\BibTeX{{\rm B\kern-.05em{\sc i\kern-.025em b}\kern-.08em
    T\kern-.1667em\lower.7ex\hbox{E}\kern-.125emX}}

\def\BibTeX{{\rm B\kern-.05em{\sc i\kern-.025em b}\kern-.08em
    T\kern-.1667em\lower.7ex\hbox{E}\kern-.125emX}}
\begin{document}

\title{Improving Infinitely Deep Bayesian Neural Networks with Nesterov’s Accelerated Gradient Method\\
\thanks{ \footnotesize \textsuperscript{*}Work done during the author's internship at the Shenzhen Institutes of Advanced Technology. $\dagger${Corresponding author: Wenqi Fang}.}
}

\author{\IEEEauthorblockN{Chenxu Yu$^*$}
\IEEEauthorblockA{\textit{Shenzhen Institutes of Advanced Technology} \\
\textit{Chinese Academy of Sciences}\\
Shenzhen, P.R.China \\
yuchenxu1024@163.com}
\and
\IEEEauthorblockN{Wenqi Fang$^\dagger$}
\IEEEauthorblockA{\textit{Shenzhen Institutes of Advanced Technology} \\
\textit{Chinese Academy of Sciences}\\
Shenzhen, P.R.China \\
wq.fang@siat.ac.cn}
}

% \and

\maketitle

\begin{abstract}
As a representative continuous-depth neural network approach, stochastic differential equation (SDE)–based Bayesian neural networks (BNNs) have attracted considerable attention due to their solid theoretical foundations and strong potential for real-world applications. However, their reliance on numerical SDE solvers inevitably incurs a large number of function evaluations (NFEs), resulting in high computational cost and occasional convergence instability. To address these challenges, we propose a Nesterov-accelerated gradient (NAG) enhanced SDE-BNN model. By integrating NAG into the SDE-BNN framework along with an NFE-dependent residual skip connection, our method accelerates convergence and substantially reduces NFEs during both training and testing. Extensive empirical results show that our model consistently outperforms conventional SDE-BNNs across various tasks, including image classification and sequence modeling, achieving lower NFEs and improved predictive accuracy.
\end{abstract}


\section{Introduction}

Neural networks (NNs) are commonly modeled as discrete sequences of layers, each consisting of an affine transformation followed by a nonlinear activation~\cite{pires2025artificial}. This formulation underpins modern deep learning and has driven success across a wide range of applications, including image classification~\cite{wang2025vision}, speech recognition~\cite{ahlawat2025automatic}, and natural language processing~\cite{ariai2025natural}. As network depth increases, NNs can be naturally interpreted as discretizations of continuous dynamical systems, providing new perspectives for theoretical analysis, architecture design, and understanding model behavior~\cite{ma2020machine}.

As the seminal continuous-depth neural network proposed by Chen et al, Neural Ordinary Differential Equations (Neural ODEs) model hidden states as solutions to an ODE, replacing discrete-layer architectures with continuous-time dynamics~\cite{chen2018neural}. This idea has spawned numerous Neural ODE variants—augmented~\cite{dupont2019augmented}, second-order~\cite{norcliffe2020second, yildiz2019ode2vae, nguyen2022improving}, control-based~\cite{kidger2020neural, asikis2022neural, wang2025pidnodes}, and physically informed~\cite{greydanus2019hamiltonian, cranmer2020lagrangian, finzi2020simplifying}—collectively enhancing the expressivity, efficiency, and versatility of continuous-depth NNs. These formulations enable Neural ODE-based approaches to excel in time-series forecasting~\cite{oh2025comprehensive}, image analysis~\cite{niu2024applications}, and dynamical system simulation~\cite{nair2025understanding}, while providing strong generalization performance and enhanced interpretability.

To account for stochastic noise, Neural SDEs extend Neural ODEs to explicitly model inherent randomness and uncertainty in the data~\cite{liu2019neural}. By embedding NNs in the deterministic drift term, stochastic diffusion term, or both, it offers a flexible framework for modeling complex dynamics that combine structured behavior with randomness, making it well-suited for stochastic process modeling, probabilistic forecasting, and noisy system simulation~\cite{jia2019neural, li2020scalable, boral2023neural, oh2024stable}.

Building upon this idea, the SDE-BNN method incorporates Bayesian uncertainty inference into the Neural SDEs framework, thereby further improving robustness and generalization~\cite{xu2022infinitely}. Although SDE-BNNs capture uncertainty through parameter distributions, their dependence on iterative SDE solvers leads to excessive function evaluations, increased computational cost, and potentially unstable convergence. To some extent, recent variants, such as partially stochastic SDE-BNN (PSDE-BNN)~\cite{calvo2024partially} and rough path theory–based SDE-BNN (RDE-BNN)~\cite{xiaoyu2025infinitely}, alleviate these challenges by introducing partial stochasticity into the SDE-BNN architecture.

Unlike prior variants, and inspired by Nesterov-accelerated NODEs~\cite{nguyen2022improving}, we incorporate the Nesterov accelerated gradient (NAG) method into the SDE-BNN framework, termed Nesterov-SDEBNN. The proposed approach improves numerical stability and computational efficiency, thereby accelerating convergence while maintaining—or even enhancing—generalization performance. In contrast to PSDE-BNN and RDE-BNN, which rely on more complex mathematical formulations, Nesterov-SDEBNN achieves these improvements by simply extending the original first-order ODEs in SDE-BNN to second-order dynamics via NAG and revising the residual connection scheme. Experiments on multiple representative open-source datasets show that Nesterov-SDEBNN significantly reduces the NFEs, accelerates convergence, and improves predictive accuracy.

The main contributions of this paper are summarized as follows:
\begin{itemize}
    \item We propose Nesterov-SDEBNN by integrating the NAG method into the original SDE-BNN framework, complemented by an NFE-dependent residual connection strategy to enhance overall performance.
    \item Compared with the original SDE-BNN, our method converges faster and achieves higher accuracy on image classification and sequence modeling tasks, while significantly reducing NFEs, thereby accelerating both training and testing in adaptive differential equation solvers.
    %\item We establish the existence of adjoint states and demonstrate that backward propagation can be efficiently performed through the adjoint differential equation, without storing intermediate states.
    %\item Compared to the original SDE-BNN, our method significantly speeds up convergence on image classification and sequence modeling tasks, while achieving improved accuracy.
    %\item Our model significantly reduces NFEs, enabling substantial acceleration of both training and inference in adaptive iterative computation of differential equations.
\end{itemize}

\section{Related Work}

\subsection{Neural Ordinary Differential Equations}
Neural ODEs parameterize continuous-time dynamics using NNs of the form:
\begin{equation}
\frac{dh_t}{dt} = f_{\theta}(h_t, t), \quad h_0 \in \mathbb{R}^d
\end{equation}
where \( h_t \) denotes the hidden state at time \( t \) and \( f: \mathbb{R}^d \times \mathbb{R} \to \mathbb{R}^d \) is a Lipschitz-continuous NN parameterized by \( \theta \)~\cite{chen2018neural}. The forward pass corresponds to numerically solving this ODE from an initial input state \( h_0 = x \), yielding an infinitely deep residual network with universal approximation capability, while enabling memory-efficient training via the adjoint sensitivity method and flexible accuracy–efficiency tradeoffs through adaptive ODE solvers~\cite{chen2018neural, dupont2019augmented, norcliffe2020second, yildiz2019ode2vae, nguyen2022improving}.

\subsection{Nesterov Neural ODEs}

To reduce the NFEs and accelerate training and testing, Nguyen et al.~\cite{nguyen2022improving} introduced the NAG method into Neural ODEs (Nesterov Neural ODE) by extending first-order ODEs to second-order dynamics.
%This acceleration significantly improves computational efficiency and convergence speed, while the proven existence of adjoint states enables memory-efficient backpropagation without storing intermediate states.
Nesterov Neural ODEs were formulated by solving the following second-order differential equation:
\begin{equation}\label{Nest}
\frac{d^2h_t}{dt^2} + \frac{3}{t} \frac{dh_t}{dt} + f_\theta(h_t, t) = 0
\end{equation}
To improve computational efficiency and mitigate numerical instability, equation (\ref{Nest}) is reformulated as an equivalent first-order system, making it more amenable to NN implementations~\cite{nguyen2022improving}:
\begin{equation}\label{Nesterov}
\left\{
\begin{aligned}
&h_t  = \sigma_f \left( t^{-\frac{3}{2}} e^{\frac{t}{2}} \right) x_t, \\
&x_t' = \sigma_f \left( m_t \right), \\
&m_t' = -m_t - \sigma_f \left( f_\theta(h_t, t) \right) - \xi h_t,
\end{aligned}
\right.
\end{equation}
Here, $x_t$ denotes an auxiliary variable associated with $h_t$, \( m_t \) represents the momentum term, $\sigma_f$ is a nonlinear activation function, and $\xi$ controls the residual connection~\cite{nguyen2022improving}. This reformulation recasts the dynamics as a first-order system that incorporates momentum, leading to accelerated convergence and serving as the basis for our methodology.

\subsection{Stochastic Differential Equations based Bayesian Neural Network}

Neural SDEs extend Neural ODEs by introducing stochastic components into the dynamics, enabling the model to capture both deterministic and stochastic behavior~\cite{liu2019neural}:
\begin{equation}
\frac{dh_t}{dt} = f_{w}(h_t, t) \, dt + g_{\theta}(h_t, t) \, dB_t
\end{equation}
where \( f_w \) and \( g_\theta \) are the drift and diffusion functions, respectively, and \( B_t \) denotes Brownian motion. This formulation allows the network to model randomness inherent in various real-world applications, such as robust pricing and hedging, irregular time series data analysis, and eddy simulation~\cite{oh2024stable, gierjatowicz2022robust, boral2023neural}.

Building on Neural SDEs, the SDE-BNN framework integrates SDEs with BNNs to capture uncertainty in both system dynamics and network parameters~\cite{xu2022infinitely, li2020scalable}. In this framework, the NN weights are modeled as stochastic processes. Therefore, training SDE-BNNs via variational inference (VI) involves solving an augmented SDE that jointly tracks the trajectories of both the weights and the network activations~\cite{xu2022infinitely}:
\begin{equation}\label{SDEBNN}
d \begin{bmatrix} w_t \\ h_t \end{bmatrix} = \begin{bmatrix} f_\phi(w_t, t) \\ f_{w_t}(h_t, t) \end{bmatrix} dt + \begin{bmatrix} g_\theta(w_t, t) \\ 0 \end{bmatrix} dB_t
\end{equation}
where \( f_{w_t} \) and \( f_{\phi} \) represent the dynamics of the weights and the activations of the network, respectively. The final hidden unit state \( h_T \ (t \in (0, T]) \) is used to parameterize the likelihood of the target output \( y \).

While SDE-BNNs effectively model uncertainty and improve robustness under noisy or incomplete observations, it suffers from a significant drawback: solving SDEs requires huge NFEs, which slows both training and testing. To mitigate this, PSDE-BNN~\cite{calvo2024partially} and RDE-BNN~\cite{xiaoyu2025infinitely} introduce partial stochasticity, striking a balance between computational cost and uncertainty modeling. Unlike these methods, our approach tackles this challenge by integrating the NAG method directly into the SDE-BNN framework, offering a simple yet effective solution.

\section{Methodology}

In this section, we present details of our Nesterov-SDEBNN approach.

\subsection{Prior Process and Approximate Posterior over Weights}

In essence, the proposed Nesterov-SDEBNN adheres to the Bayesian framework and thus admits well-defined prior and posterior distributions.

To obtain a simple prior with bounded marginal variance in the long-time limit, we adopt an Ornstein–Uhlenbeck (OU) process as the prior over the weights, defined by an SDE with the following drift and diffusion terms:
\begin{equation}
f_p(w_t, t) = -w_t, \quad g(w_t, t) = \sigma I_d,
\end{equation}
where \( \sigma \) is a hyperparameter and $I_d$ is a $d \times d$ identity matrix.

For the posterior, we seek a posterior approximation that can capture non-Gaussian, non-factorized marginals, achieved by parameterizing its dynamics with a NN whose capacity can be scaled as needed. Accordingly, we implicitly define the approximate posterior over the weights via an SDE with a learned drift:
\begin{equation}
f_q(w_t, t, \phi) = \mathrm{NN}_{\phi}(w_t, t) - f_p(w_t, t),
\end{equation}
where the posterior drift, \( f_q \), is parameterized by a small NN with parameters \( \phi \), while the diffusion terms are kept the same as the prior.

\subsection{Evaluation of the Network}

\begin{figure}[t]
    \centering
    \begin{minipage}[b]{0.45\columnwidth}
         \centering
         \includegraphics[width=\linewidth]{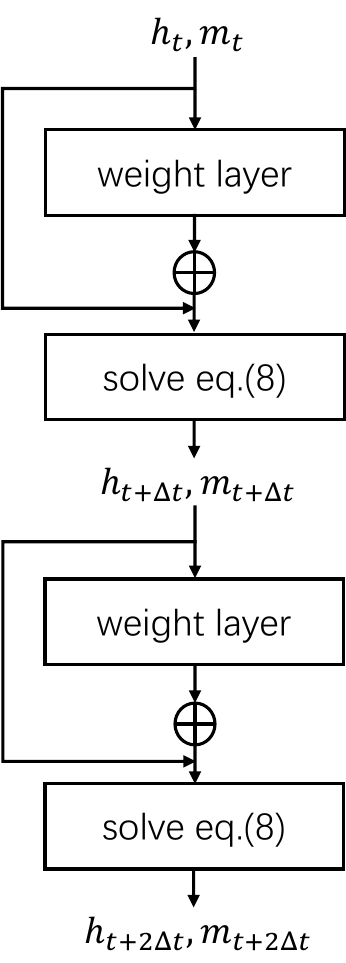}
         \subcaption{}
     \end{minipage}%
     \hfill
     \begin{minipage}[b]{0.45\columnwidth}
         \centering
         \includegraphics[width=\linewidth]{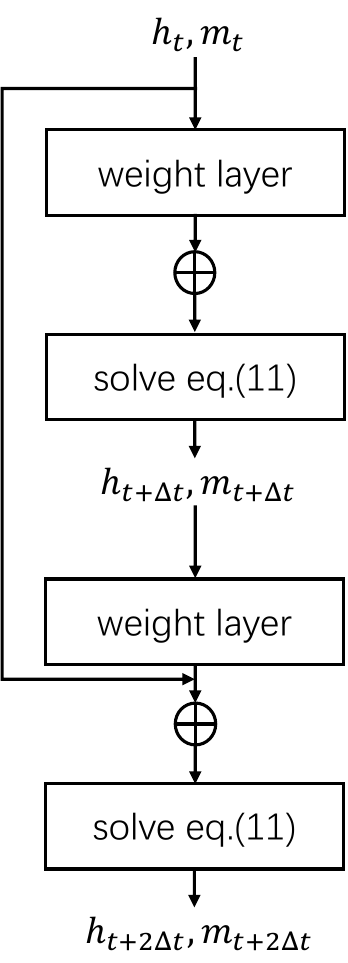}
         \subcaption{}
     \end{minipage}
     \vspace{-5pt}
    \caption{Comparison of mentioned residual mechanisms:
    \textbf{$(a)$} Direct residual connection for $m_t$ in eq (\ref{sde}).
    \textbf{$(b)$} Proposed $\text{NFE}_f$-dependent residual skip connection for $m_t$ in eq (\ref{final}).
}
    \label{fig:structure_comparison}
\end{figure}

By incorporating the NAG method into the SDE-BNN framework, we can easily derive the modified dynamics by combining equations (\ref{Nesterov}) and~(\ref{SDEBNN}), expressed in terms of the joint state \( H_t := (x_t, m_t, w_t) \):
\begin{equation}\label{sde}
\begin{aligned}
&h_t  = \sigma_f \left( t^{-\frac{3}{2}} e^{\frac{t}{2}} \right) x_t, \\
&d H_t = \begin{bmatrix} 
\sigma_f(m_t) \\ 
-m_t-\sigma_f (f_{w_t}(h_t, t))-\xi h_t \\ 
f_{\phi}(w_t, t) 
\end{bmatrix} dt 
+ \begin{bmatrix} 
0 \\ 0 \\ g(w_t, t)  
\end{bmatrix} dB_t.
\end{aligned}
\end{equation}
%\paragraph{Limitation of the naive shortcut}
Notably, the residual term $\xi h_t$ in equation (\ref{sde}) is added to evaluate the drift of $m_t$. Due to the time-stepping nature of SDE solvers, each increment of $\text{NFE}_f$ (denotes the number of drift function evaluations) advances the joint state from $H_t$ to $H_{t+\Delta t}$. Specifically, to evolve the state from $m_t$ to $m_{t+\Delta t}$, the feature $h_t$ is first processed by the function $f_{w_t}$ with weight $w_t$ and activation $\sigma_f$, and then combined with the residual $\xi h_t$ and the current state $m_t$. After solving equation (\ref{sde}), we obtain $h_{t+\Delta t}$ and $m_{t+\Delta t}$. These processes are repeated iteratively until the final time $T$,  as illustrated in subfigure~(a) of Figure~\ref{fig:structure_comparison}. Therefore, this residual mechanism establishes explicit link between representation at successive time steps (e.g., $m_t \rightarrow m_{t+\Delta t} \rightarrow m_{t+2\Delta t}$) rather than within a single time step. However, these settings result only in a direct residual connection, rather than the residual skip connection defined in the original formulation~\cite{he2016deep}. This discrepancy may reduce feature reuse and constrain the effectiveness of residual learning in the SDEBNN settings.

To better align with classic residual learning, rather than injecting residual at every drift evaluation, we propose an $\text{NFE}_f$-dependent residual skip connection mechanism. Specifically, we modify the drift of $m_t$ as:
\begin{equation}\label{mmm}
    \frac{d m_t}{dt} = -m_t - \sigma_f \left( f_{w_t}(h_t, t) + \epsilon \xi h_{\text{temp}} \right),
\end{equation}
where the control variables $\epsilon$ and $h_{\text{temp}}$ (initialized as input $x$, and retained from the previous value when $\text{NFE}_f$ is odd) depend on $\text{NFE}_f$, defined as:
\begin{equation}\label{NFE}
\epsilon =
\begin{cases}
1, & \text{if } \text{NFE}_f \text{ is odd} \\
0, & \text{if } \text{NFE}_f \text{ is even}
\end{cases},
h_{\text{temp}} = h_t \ \text{if } \text{NFE}_f \text{ is even}.
\end{equation}
Consequently, the final formulation of our Nesterov-SDEBNN method becomes:
\begin{equation}\label{final}
\begin{aligned}
&h_t  = \sigma_f \left( t^{-\frac{3}{2}} e^{\frac{t}{2}} \right) x_t, \\
&d H_t = \begin{bmatrix} 
\sigma_f(m_t) \\  
-m_t - \sigma_f \left( f_{w_t}(h_t, t) + \epsilon\xi h_{\text{temp}} \right) \\ 
f_{\phi}(w_t, t) 
\end{bmatrix} dt 
+ \begin{bmatrix} 
0 \\ 0 \\ g(w_t, t) 
\end{bmatrix} dB_t.
\end{aligned}
\end{equation}
%where $g$ is shorthand for $g(w_t, t)$.

As described in equations (\ref{mmm}) and (\ref{NFE}), the proposed $\text{NFE}_f$-dependent scheme adheres to the spirit of residual skip connections, as shown in subfigure~(b) of Figure~\ref{fig:structure_comparison}. When $\text{NFE}_f$ is odd, the feature $h_t$ is cached for use at time $t+\Delta t$; when $\text{NFE}_f$ becomes even $(\neq 0)$, the cached feature is injected into the drift at $t+\Delta t$ to help produce $m_{t+2\Delta t}$. This mechanism creates a skip connection across two consecutive drift evaluations, reusing cached representations to improve information flow without adding parameters. Meanwhile, this formulation allows the state \(m_{t+2\Delta t}\) to depend explicitly on both \(h_t\) and \(h_{t+\Delta t}\), rather than only on \(h_{t+\Delta t}\) as in equation (\ref{sde}), which may help mitigate vanishing gradient issues commonly encountered in SDE solvers.

%\paragraph{Monte Carlo evaluation}
To compute $h_t$ given $x_0 =h_0 = x$, we marginalize over weight trajectories via Monte Carlo sampling. Specifically, we sample paths $\{w_t\}$ from the posterior process, and for each path solve equation (\ref{final}) to obtain $\{h_t\}$. The learnable parameters include the initial weights $w_0$ and the drift parameters $\phi$.

\subsection{Output Likelihood}

Predictions are generated from the final hidden representation \( h_T \), which directly parameterizes the output likelihood: \( \log p(y | x, w) = \log p(y | h_T) \). In practice, this likelihood can be chosen as a Gaussian distribution for regression tasks or a categorical distribution for classification.

\subsection{Training objective}

To train our Nesterov-SDEBNN model, we adopt the VI framework, in which the loss function is derived from the evidence lower bound (ELBO) by Monte Carlo sampling~\cite{xu2022infinitely}, defined as:
\begin{equation}
\mathcal{L}_{\text{ELBO}}(\phi) = \mathbb{E}_{q_{\phi}(w)} \left[ \log p(y|x,w) - \int_0^T \frac{1}{2} \|u(w_t, t, \phi)\|_2^2 \, dt \right]
\end{equation}
where \( u(w_t, t, \phi) = g(w_t, t)^{-1} \left[ f_q(w_t, t, \phi) - f_p(w_t, t) \right] \), $q_{\phi}(w)$ denotes the posterior distribution of $w$ for short. 
%The first term represents the data likelihood, the second is the KL divergence between prior and posterior dynamics, and 
The sampled weights, hidden activations, and training objective can be computed simultaneously via a single SDE solver call~\cite{xu2022infinitely}.

\section{Experiments}

In this section, we empirically evaluate the proposed Nesterov-SDEBNN against the baseline SDE-BNN on a diverse set of benchmark tasks, including toy regression, image classification, and dynamical simulation. These benchmarks cover multiple data modalities, ranging from images to time-series data. All experiments were implemented using PyTorch and conducted on a remote server equipped with an NVIDIA A100 GPU (40 GB memory) and an Intel Xeon Gold 5320 CPU (26 cores). Additional details on training procedures, model configurations, and datasets are provided in Table~\ref{tab:hparam settings}. The block configurations follow those of the original SDE-BNN approach~\cite{xu2022infinitely}.

\begin{table*}[t]
\caption{The following hyperparameters are used for each evaluation method corresponding to the experimental results reported in the paper.}
\centering
\setlength{\tabcolsep}{0.8em}
% \resizebox{1\textwidth}{!}{
\begin{tabular}{@{} l l l l l l l @{}}\toprule
  & \multicolumn{1}{c}{} &
  & \multicolumn{1}{c}{Experiments} &
  & \multicolumn{1}{c}{} & \\
  % & \multicolumn{1}{c}{} & \\
\cmidrule(l){3-6}
Model & Hyper-parameter & 1D Regression
& MNIST~\cite{deng2012mnist} & CIFAR-10~\cite{krizhevsky2014cifar} & Walker2D~\cite{todorov2012physics} \\
\cmidrule(r){1-2} \cmidrule(l){3-6}
\textbf{SDE-BNN (fixed step)}
\\ \textbf{and Nesterov-SDE-BNN (fixed step)}
 & Augment dim. & 2 & 2 & 2 & 0 \\
 & \# blocks & 1 & 1 & 1-1-1 & 1 \\
 & Diffusion $\sigma$ & 0.2 & 0.1 & 0.1 & 0.1 \\
 & KL coef. & 0.0 & 1e-5 & 100 & 1 \\
 & Learning Rate & 1e-3 & 1e-3 & 3e-4 & \{0:1e-3, 50:3e-3\} \\
 & \# Solver Steps & 20 & 20 & 20 & 50 \\
 & Batch Size & 50 & 128 & 128 & 256 \\
 & Activation & swish & swish & mish & tanh \\
 & Epochs & 1000 & 100 & 500 & 500 \\
 & Drift $f_x$ dim. & 32 & 32 & 64 & 24 \\
 & Drift $f_w$ dim. & 32 & 1-64-1 & 2-32-2 & 1-32-1 \\
 & \# Posterior Samples & 10 & 1 & 1 & 1 \\
 & solver & midpoint & midpoint & midpoint & midpoint \\
\cmidrule(r){1-2} \cmidrule(l){3-6}
\textbf{SDE-BNN (adaptive)}
\\ \textbf{and Nesterov-SDE-BNN (adaptive)}
 & $<$SDE-BNN$>$ & - & $<$SDE-BNN$>$ & $<$SDE-BNN$>$ & $<$SDE-BNN$>$ \\
 & atol & - & 1e-3 & 5e-3 & 5e-3 \\
 & rtol & - & 1e-3 & 5e-3 & 5e-3 \\
 & Epochs & - & 100 & 300 & 500 \\
 & Activation & - & swish & swish & tanh \\
 & solver & midpoint & midpoint & midpoint & midpoint \\
\bottomrule
\end{tabular}
\label{tab:hparam settings}
% }
\end{table*}

\subsection{Toy 1D Regression}
\begin{figure}
    \centering
    \includegraphics[width=0.9\linewidth]{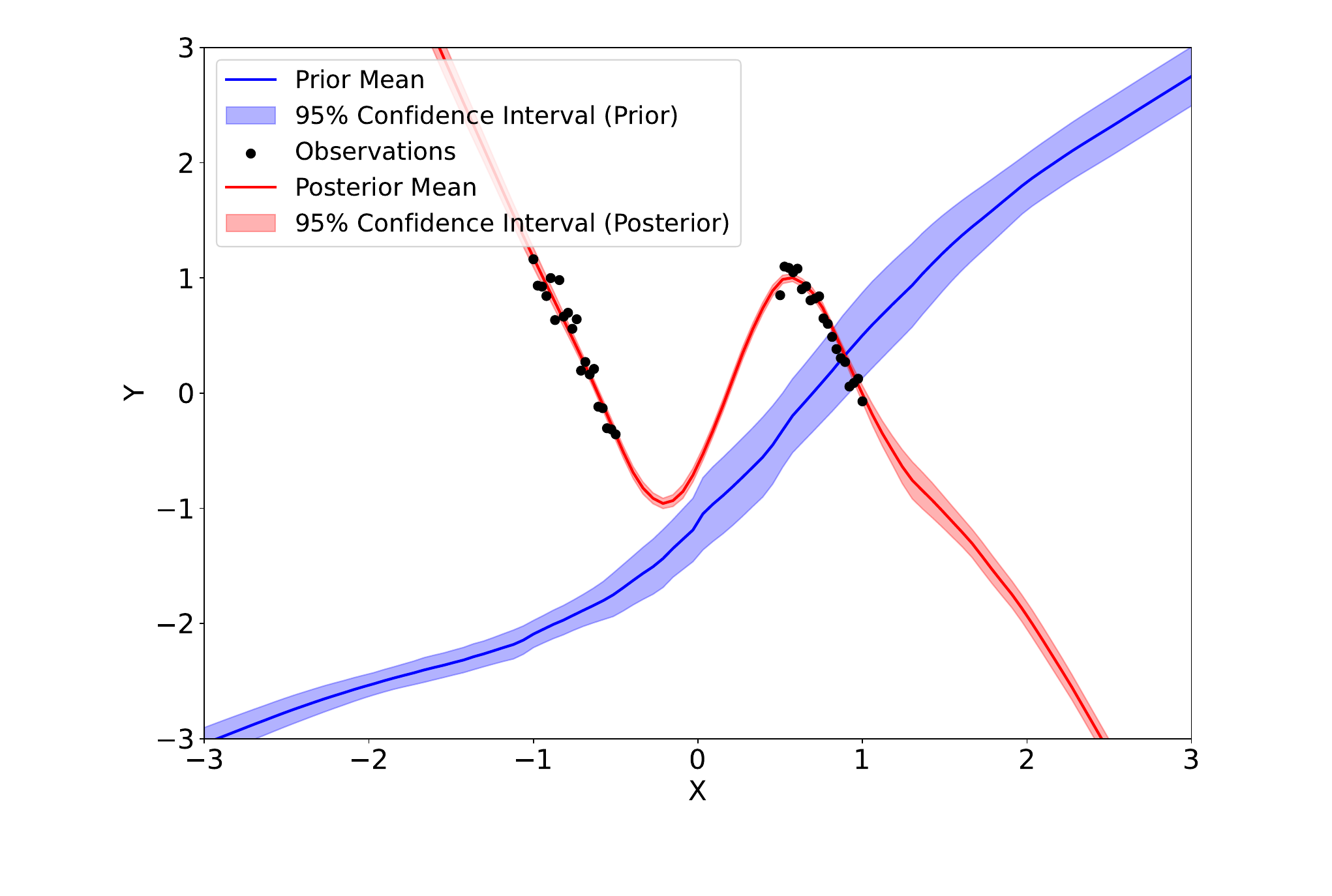}
    \vspace{-10pt}
    \caption{Predictive prior and posterior of Nesterov-SDEBNN on a non-monotonic toy dataset. The blue and red shaded regions denote the 95\% confidence intervals of the prior and posterior, respectively, while the solid lines represent their corresponding mean predictions.
    }
    \label{fig:toypriorposterior}
\end{figure}

We first validate the capabilities of Nesterov-SDEBNN on a simple 1D regression problem. To model non-monotonic functions, we augment the state space by adding two additional dimensions initialized to zero. As shown in Figure~\ref{fig:toypriorposterior}, our model retains the expressive posterior predictive power of SDE-BNN on synthetic non-monotonic noisy 1D data, demonstrating its effectiveness on this task.on this simple fitting task.

\subsection{Image Classification}

\begin{figure}
    \centering
    \begin{minipage}[b]{0.48\columnwidth}
         \centering
         \includegraphics[width=\linewidth]{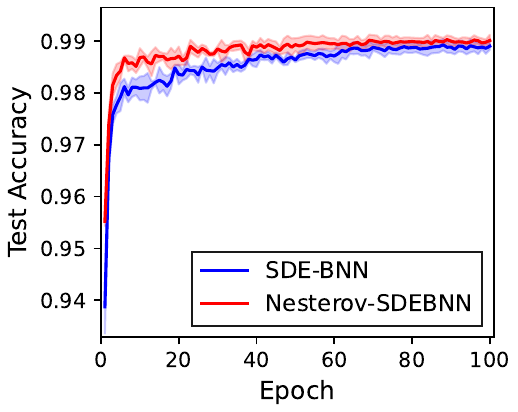}
         % \caption{MNIST Accuracy}
         \label{fig:mnist}
     \end{minipage}%
     \hfill
     \begin{minipage}[b]{0.48\columnwidth}
         \centering
         \includegraphics[width=\linewidth]{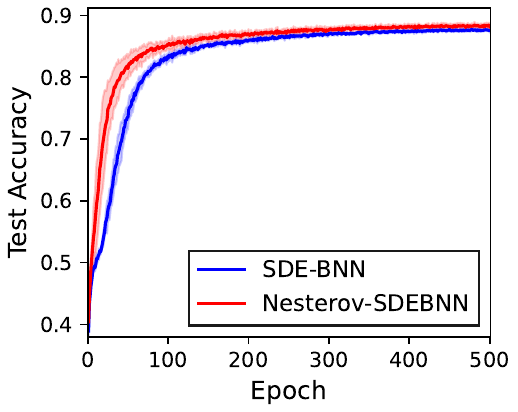}
         % \caption{CIFAR-10 Accuracy}
         \label{fig:cifar}
     \end{minipage}
     \vspace{-15pt}
    \caption{Comparison of test accuracy between SDE-BNN and Nesterov-SDEBNN: (Left) MNIST (Right) CIFAR-10.}
    \label{fig:accuracy_comparison}
\end{figure}

\begin{figure}
    \centering
    \begin{minipage}[b]{0.48\columnwidth}
         \centering
         \includegraphics[width=\linewidth]{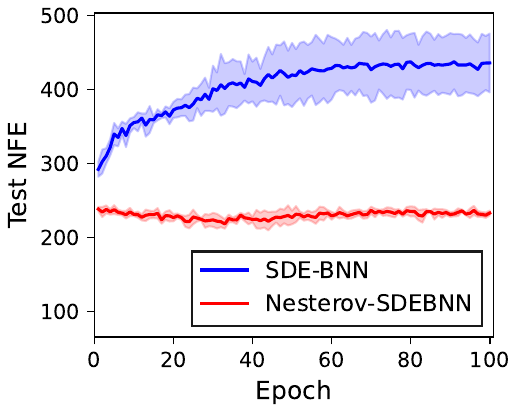}
         % \caption{MNIST Test NFE}
         \label{fig:mnist_nfe}
     \end{minipage}%
     \hfill
     \begin{minipage}[b]{0.48\columnwidth}
         \includegraphics[width=\linewidth]{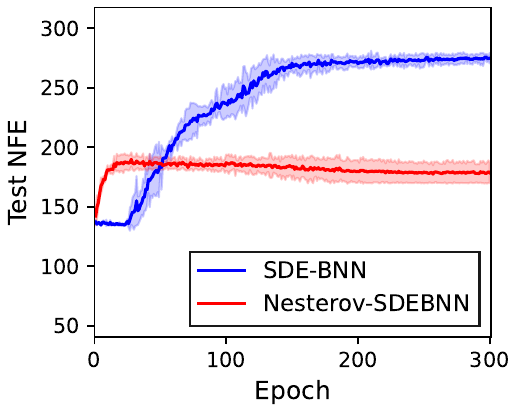}
         % \caption{CIFAR-10 Test NFE}
         \label{fig:cifar_nfe}
     \end{minipage}%
     \vspace{-15pt}
    \caption{Comparison of test NFEs between SDE-BNN and Nesterov-SDEBNN: (Left) MNIST (Right) CIFAR-10.}
    \label{fig:nfe_comparison}
\end{figure}

For image classification, we followed the SDE-BNN setup, using convolutional networks to model instantaneous hidden-state changes. We benchmarked Nesterov-SDEBNN against SDE-BNN on MNIST and CIFAR-10 in terms of accuracy, negative log-likelihood (NLL), area under the curve (AUC), and NFEs, with results reported in Table~\ref{tab1:img_results}, Figures~\ref{fig:accuracy_comparison} and \ref{fig:nfe_comparison}.

\begin{table*}\centering
\renewcommand{\arraystretch}{1.1}
\setlength{\tabcolsep}{0.8em}
\caption{Quantitative evaluation on MNIST and CIFAR-10. Comparison of SDE-BNN and Nesterov-SDEBNN performance (Accuracy, AUC, and NLL) using fixed-step and adaptive-step integration.}
% \resizebox{\textwidth}{!}{
\begin{tabular}{@{} l c c c c c c c @{}}\toprule
  & \multicolumn{3}{c}{\textbf{\textsc{MNIST}}} & & \multicolumn{3}{c}{\textbf{\textsc{CIFAR-10}}} \\
\cmidrule(lr){2-4} \cmidrule(l){6-8}
Model & \tcen{Accuracy (\%)$\boldsymbol{\uparrow}$} & \tcen{AUC (\%)$\boldsymbol{\uparrow}$} & \tcen{NLL ($\times 10^{-2}$)$\boldsymbol{\downarrow}$} & & \tcen{Accuracy (\%)$\boldsymbol{\uparrow}$} & \tcen{AUC (\%)$\boldsymbol{\uparrow}$} & \tcen{NLL ($\times 10^{-1}$)$\boldsymbol{\downarrow}$} \\
\cmidrule(r){1-1}\cmidrule(lr){2-4} \cmidrule(l){6-8}
SDE-BNN(fixed step) & 98.90 $\pm$ 0.07 & 97.59 $\pm$ 0.04 & 14.37 $\pm$ 0.89 & & 87.56 $\pm$ 0.36 & 83.05 $\pm$ 0.29 & 9.89 $\pm$ 0.22 \\
Nesterov-SDEBNN(fixed step) & \textbf{99.04} $\pm$ 0.12 & \textbf{97.87} $\pm$ 0.06 & \textbf{7.37} $\pm$ 0.65 & & \textbf{88.36} $\pm$ 0.24 & \textbf{85.61} $\pm$ 0.45 & \textbf{5.97} $\pm$ 0.11 \\

\cmidrule(r){1-1}\cmidrule(lr){2-4} \cmidrule(l){6-8}
SDE-BNN(adaptive step) & 98.87 $\pm$ 0.12 & 97.57 $\pm$ 0.03 & 13.85 $\pm$ 1.26 & & 85.87 $\pm$ 0.28 & 78.42 $\pm$ 0.60 & 6.77 $\pm$ 0.09 \\
Nesterov-SDEBNN(adaptive step) & \textbf{99.04} $\pm$ 0.03 & \textbf{97.88} $\pm$ 0.04 & \textbf{7.15} $\pm$ 0.34 & & \textbf{86.99} $\pm$ 0.23 & \textbf{83.94} $\pm$ 0.32 & \textbf{5.32} $\pm$ 0.10 \\
\bottomrule
\end{tabular}
% }
\label{tab1:img_results}
\end{table*}

Table~\ref{tab1:img_results} summarizes the performance of SDE-BNN and Nesterov-SDEBNN on MNIST and CIFAR-10 datasets under both fixed-step and adaptive integration schemes. Across all settings, Nesterov-SDEBNN consistently achieves slightly higher accuracy and AUC than the standard SDE-BNN. Notably, the NLL values are significantly lower for Nesterov-SDEBNN, indicating improved predictive uncertainty. On MNIST, the fixed-step Nesterov-SDEBNN reaches 99.04\% accuracy and $7.37 \times 10^{-2}$ NLL, compared to 98.90\% and $14.37 \times 10^{-2}$ for SDE-BNN. On CIFAR-10, similar trends are observed, with Nesterov-SDEBNN improving both accuracy (88.36\% vs. 87.60\%), AUC (85.61\% vs. 83.05\%), and NLL ($5.97 \times 10^{-1}$ vs. $9.89 \times 10^{-1}$) under fixed-step conditions. In addition, in adaptive integration scenarios, Nesterov-SDEBNN consistently reduces NLL while simultaneously improving predictive accuracy. This dual improvement underscores the model's robustness in both discriminative performance and uncertainty calibration, regardless of the underlying step-size scheme or dataset complexity.

Figure \ref{fig:accuracy_comparison} compares the test accuracy of the two models. While both achieve similar final accuracy, the Nesterov-SDEBNN curve consistently lies above that of SDE-BNN throughout testing, resulting in a larger AUC value. This indicates that Nesterov-SDEBNN converges faster and achieves higher accuracy in the early stages of training on both the MNIST and CIFAR-10 datasets.

Furthermore, we compares the test NFEs between SDE-BNN and Nesterov-SDEBNN on MNIST and CIFAR-10, as shown in Figure~\ref{fig:nfe_comparison}. On both datasets, Nesterov-SDEBNN consistently requires significantly fewer NFEs than SDE-BNN throughout training, indicating more efficient computation. For MNIST, SDE-BNN’s NFE increases steadily and fluctuates around 400, whereas Nesterov-SDEBNN remains stable near 240. On CIFAR-10, SDE-BNN’s NFE rises sharply after around 50 epochs and plateaus near 270, while Nesterov-SDEBNN stabilizes around 170, demonstrating faster convergence and lower computational cost compared to SDE-BNN.

These results demonstrate that Nesterov-SDEBNN delivers superior classification performance while significantly reducing computational overhead, making it a highly efficient solution for image classification.

\subsection{Walker2D Kinematic Simulation}
To evaluate the model's capacity for capturing long-term dependencies \cite{lechner2006learning}, we applied Nesterov-SDEBNN to Walker2D kinematic simulation data~\cite{todorov2012physics}. Building on the ODE-RNN framework \cite{rubanova2019latent}, we benchmarked our approach against the standard SDE-BNN.

\begin{figure}
    \centering
    \begin{minipage}[b]{0.48\columnwidth}
         \centering
         \includegraphics[width=\linewidth]{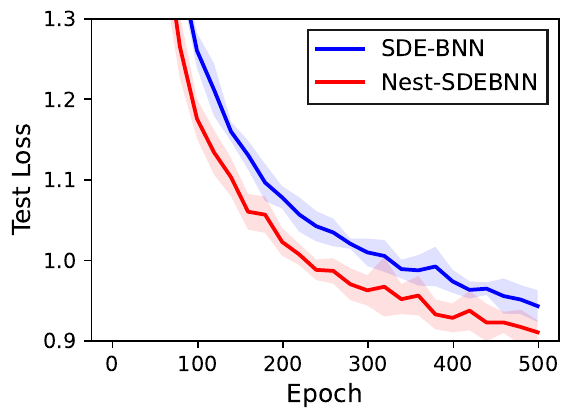}
         \label{fig:walker2d_test_loss_nonadaptive}
     \end{minipage}%
     \hfill
     \begin{minipage}[b]{0.48\columnwidth}
         \centering
         \includegraphics[width=\linewidth]{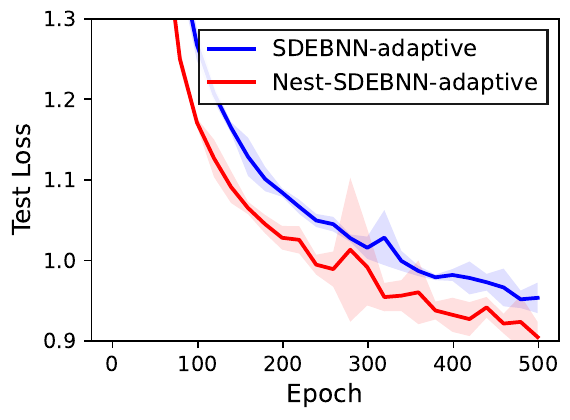}
         \label{fig:walker2d_test_loss_adaptive}
     \end{minipage}
     \vspace{-15pt}
    \caption{Walker2D test loss performance. Nesterov-SDEBNN vs. SDE-BNN under fixed-step (Left) and adaptive-step (Right) solver configurations.}
    \label{fig:walker2d_test_loss}
\end{figure}

As shown in Figure~\ref{fig:walker2d_test_loss}, Nesterov-SDEBNN reduces loss faster and achieves lower final error rates than standard counterparts. Despite slightly higher volatility and oscillations during testing, it maintains superior performance throughout the 500 epochs, with its mean loss curve generally below the baseline.

\begin{figure}
    \centering
    \begin{minipage}[b]{0.48\columnwidth}
         \centering
         \includegraphics[width=\linewidth]{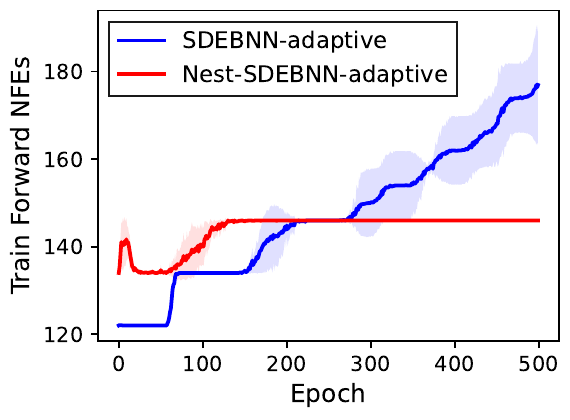}
         \label{fig:walker2d_train_forward_nfe}
     \end{minipage}%
     \hfill
     \begin{minipage}[b]{0.48\columnwidth}
         \centering
         \includegraphics[width=\linewidth]{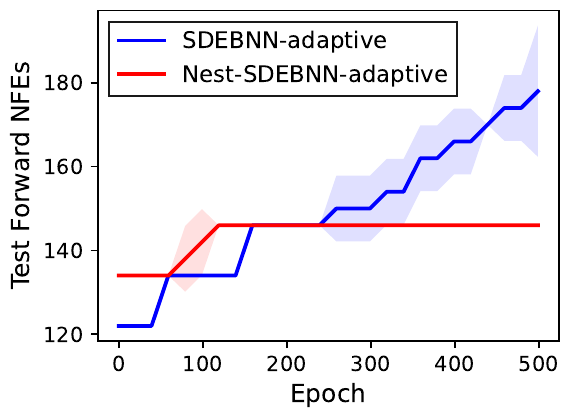}
         \label{fig:walker2d_test_forward_nfe}
     \end{minipage}
     \vspace{-15pt}
    \caption{Forward-pass NFEs on the Walker2D kinematic simulation. Comparison between SDE-BNN and Nesterov-SDEBNN during (Left) training and (Right) testing using adaptive-step solver.}
    \label{fig:walker2d_forward_nfe}
\end{figure}

Furthermore, as illustrated in Figure~\ref{fig:walker2d_forward_nfe}, Nesterov-SDEBNN demonstrates superior efficiency during both training and testing phases. While it initially incurs slightly higher NFEs, it quickly stabilizes around 145, maintaining a consistent computational cost. In contrast, standard SDE-BNN shows a stepwise increase in complexity, reaching 180 NFEs, about 24\% higher than ours. These results indicate that integrating Nesterov-like dynamics into the SDE-BNN framework effectively limits solver complexity, resulting in more predictable and efficient inference than the standard adaptive approach.

Overall, these results provide further compelling empirical evidence that our approach outperforms the SDE-BNN baseline, underscoring its effectiveness and practical potential for time-series modeling.

% \begin{figure}
%     \centering
%     \includegraphics[width=\linewidth]{plots/walker2d_longterm.pdf}
%     \caption{
%     L2-norm of the backpropagated states for SDE-BNN-RNN and Nest-SDE-BNN-RNN under both fixed-step and adaptive-step settings, obtained by backpropagation from the final time stamp. The term (T - t) denotes the temporal gap between the final time (T) and an intermediate time (t). As (T - t) increases, Nest-SDE-BNN-RNN consistently exhibits a slower decay than SDE-BNN-RNN in both fixed-step and adaptive-step cases, indicating its superior ability to mitigate the vanishing gradient problem.
%     }
%     \label{fig:walker2d_longterm}
% \end{figure}

\section{Conclusion}
In this paper, we introduced Nesterov-SDEBNN, an advancement of the SDE-BNN framework that integrates Nesterov accelerated gradient principles with an NFE-dependent residual scheme. This architecture significantly curtails the computational overhead of training and testing by reducing the required number of function evaluations. Empirical results show that our approach converges faster and achieves higher accuracy on various tasks such as image classification and time-series modeling, indicating its stronger capacity to capture complex patterns than standard SDE-BNNs. Our future work will assess the framework’s scalability and robustness on larger, more complex datasets and explore extensions to examine its potential in practical applications.

% \section*{CRediT authorship contribution statement}
% Chenxu Yu: Writing – original draft, Software, Methodology.
% Wenqi Fang: Writing – review \& editing, Validation, Supervision, Investigation, Funding.

% \section*{Declaration of competing interest}
% The authors declare that there are no conflicts of interest regarding the publication of this paper.

\section*{Acknowledgments}
W.Fang was supported by the National Natural Science Foundation of China (NSFC) under Grant No.12401676.

\bibliographystyle{unsrt}
\bibliography{references.bib}

\end{document}